# WildfireGenome: Interpretable Machine Learning Reveals Local Drivers of Wildfire Risk and Their Cross-County Variation

Chenyue Liu[1*], Ali Mostafavi[2]

[1]Ph.D. Candidate, Urban Resilience.AI Lab, Zachry Department of Civil and Environmental Engineering, Texas A&M University, College Station, United States; e-mail: liuchenyue@tamu.edu

[2]Professor, Urban Resilience.AI Lab Zachry Department of Civil and Environmental Engineering, Texas A&M University, College Station, United States; e-mail: amostafavi@civil.tamu.edu

## Abstract

Current wildfire risk assessment methods rely on coarse hazard maps and black-box machine learning models that optimize for regional accuracy while sacrificing interpretability at the decision scale where land-use and infrastructure choices occur. These approaches rarely integrate multiple risk indicators into coherent frameworks, fail to expose nonlinear thresholds in key drivers, and lack evidence on cross-regional portability. This study presents WildfireGenome, an interpretable framework that addresses these limitations through three integrated components: fusion of seven federal wildfire indicators into a sign-aligned, PCA-based composite risk label at H3 Level-8 grid resolution; Random Forest models for local risk classification; and SHAP and ICE/PDP analyses that reveal county-specific, nonlinear driver relationships. Evaluation across seven ecologically diverse United States counties (Butte, Sonoma, and El Dorado in California; Jackson in Oregon; Boulder and Larimer in Colorado; Coconino in Arizona) demonstrates strong within-county performance, with accuracy ranging from 0.755 to 0.878 and Quadratic Weighted Kappa reaching 0.951. Principal components capture 87 to 94 percent of indicator variance, validating the composite labeling approach. Cross-county transferability analysis reveals critical operational insights: models transfer effectively between ecologically similar regions such as Colorado's Front Range counties but fail dramatically between dissimilar contexts, with some transfers performing below chance levels. Explanatory analyses consistently identify needleleaf forest coverage and elevation as dominant risk drivers, with ICE curves revealing actionable thresholds where risk escalates sharply at approximately 30 to 40 percent needleleaf coverage. These county-specific nonlinearities translate directly into targeted interventions for vegetation management, zoning decisions, and infrastructure planning. The framework provides operational guidance for model deployment, establishing when existing models serve as credible screening tools versus when local refitting becomes essential. WildfireGenome advances wildfire risk assessment from opaque regional products to transparent, decision-scale analytics that support targeted adaptation strategies and operational planning where actual management decisions occur.

# Introduction

Wildfires are becoming increasingly frequent, severe, and spatially extensive across many regions of the world, driven by the combined effects of climate change, land-use transformations, and fuel accumulation [1, 2]. In the western United States, the number of large fires and total burned area has surged over the past several decades, placing unprecedented strain on emergency response systems, infrastructure, and vulnerable populations [3-5]. These compounding hazards represent not only ecological disturbances but also social crises, exacerbating existing inequities in housing, mobility, and public health. As wildfires increasingly intersect with the built environment, there is growing demand for predictive frameworks that can not only estimate risk levels but also explain the underlying drivers and guide spatially targeted adaptation strategies.

Traditional wildfire risk assessments often rely on coarse-scale hazard maps, fire behavior simulators, or empirical indices derived from historical fire records [6, 7]. While these tools offer valuable historical context and support broad hazard zoning, they often obscure localized variation in fire susceptibility arising from fine-scale interactions among vegetation structure, topography, and climatic gradients [8, 9]. Moreover, many existing approaches function as black-box models—producing predictions without clear attribution—thereby limiting their utility for climate adaptation planning, risk communication, or actionable policy design. Recent advances in machine learning (ML) and deep learning (DL) have improved wildfire prediction capabilities by capturing nonlinear and high-dimensional interactions across diverse environmental inputs [10-12]. Models such as random forests, support vector machines, and deep neural networks have been employed to predict fire occurrence, spread, or severity. However, most of these applications prioritize predictive accuracy over interpretability, and are typically implemented at regional or national scales [13, 14]. As a result, they offer limited insight into the localized feature interactions that influence fire risk—precisely the level at which land-use decisions, infrastructure investment, and managed retreat policies are made.

To address these limitations, we introduce a high-resolution, interpretable modeling framework—termed the Wildfire Genome (Figure 1)—that integrates composite risk classification, supervised ML, and post hoc explainability tools to characterize spatial wildfire susceptibility across heterogeneous landscapes. Building on recent advances in flood modeling, including the FloodGenome framework which demonstrated strong cross-region transferability, this study extends the genome-style approach to wildfire hazards. However, unlike floods, where hydrological processes exhibit greater spatial consistency, wildfire dynamics are governed by localized ecological, topographic, and fuel-related factors. These differences limit model portability and underscore the need for place-based modeling that accounts for spatial heterogeneity in fire drivers. As such, the wildfire genome framework emphasizes individualized attribution, nonlinear response analysis, and localized policy relevance.

We apply this framework to seven wildfire-prone U.S. counties, selected for their ecological and climatic diversity. Our modeling pipeline includes: (1) unsupervised composite risk classification using principal component analysis (PCA); (2) supervised risk prediction via random forest classifiers; (3) SHAP-based global feature attribution to identify dominant risk drivers and their directionality; and (4) Individual Conditional Expectation (ICE) and Partial Dependence Plot (PDP) analyses to examine fine-grained, nonlinear responses across individual samples. We evaluate both within-county model performance and cross-county transferability to assess the generalizability of wildfire risk patterns across space. While our models achieve strong predictive accuracy at the local level, transferability across counties remains limited—highlighting fundamental contrasts with prior findings in flood risk modeling and reinforcing the importance of context-specific adaptation strategies.

By combining interpretable machine learning with high-resolution spatial data, this study contributes to both wildfire risk science and the operationalization of climate adaptation. The wildfire genome framework offers a scalable yet locally grounded approach for diagnosing hazard exposure, identifying sensitive ecological thresholds, and informing spatially explicit migration and land-use decisions. It also advances the frontier of multi-hazard analytics by demonstrating how individualized model interpretation—through tools such as SHAP and ICE—can move beyond aggregate risk maps toward anticipatory, equity-centered resilience planning. While the limited cross-county transferability cautions against overgeneralization, the individualized insights offered by the wildfire genome can support context-aware governance of climate-induced displacement, wildfire mitigation, and infrastructure vulnerability.

Accordingly, this study addresses the following research questions:

1. *Within-county drivers and thresholds.* Which environmental predictors most strongly influence whether an H3 Level-8 grid cell falls into the "Very High" wildfire risk class, and what nonlinear thresholds or saturation effects characterize these relationships at decision scale? This analysis employs expected-level SHAP values to determine global importance and direction, while ICE and PDP curves reveal local response patterns.
2. *Predictive performance at decision scale.* How accurately can Random Forest models predict the PCA-derived, sign-aligned composite wildfire risk labels generated from seven federal indicators within each county? Performance is evaluated using Accuracy, Macro-F1, and Quadratic Weighted Kappa metrics on held-out H3 cells to ensure robust validation.
3. *Portability and explanation coherence.* To what extent do models trained in one county generalize to others with distinct topographic, vegetative, and climatic profiles? This question examines how differences in driver profiles revealed through SHAP rankings and directions, combined with response shape variations shown in ICE and PDP curves, explain the success or failure of cross-county transferability.

Addressing these research questions achieves three specific objectives. First, it fuses seven federal risk components into a sign-aligned, PCA-based composite label at H3 Level-8 resolution, creating a unified risk metric. Second, it employs Random Forest models to learn local risk classes, then applies SHAP and ICE/PDP techniques to reveal county-specific, nonlinear thresholds in key drivers, making the model's decision process transparent. Third, it quantifies both within-county performance and cross-county transferability across seven ecologically diverse U.S. counties, establishing empirically when local models are necessary and how explanatory insights transfer between regions. This approach addresses a critical need in wildfire risk assessment. By coupling high-resolution prediction with transparent and portable explanations, WildfireGenome provides policy-ready evidence for zoning decisions, vegetation and fuel management strategies, and climate-resilient migration planning. The framework thus advances wildfire risk assessment from coarse, black-box mapping to interpretable analytics that operate at the scale where actual land-use and infrastructure decisions are made.

# Background

***Evolution of wildfire risk assessment approaches.*** Spatial wildfire risk assessment has traditionally relied on coarse hazard products, fire behavior simulators, and empirical danger indices synthesized from long-term fire records and expert fuel models. While these approaches have provided the foundation for regional hazard zoning and national assessments, they frequently smooth over fine-scale heterogeneity arising from complex interactions among vegetation structure, topography, and climatic gradients. This limitation becomes critical at the precise scales where land-use planning, mitigation strategies, and evacuation decisions must be implemented [6,7]. The urgency for decision-scale evidence has intensified as wildfires increase in frequency, severity, and spatial extent across the western United States, placing unprecedented pressure on infrastructure, emergency response systems, and vulnerable populations [1–5]. Although traditional products remain essential for situational awareness, their limited spatial granularity and opaque driver attribution constrain their utility for targeted, place-based adaptation strategies.

***Remote sensing contributions and constraints.*** Contemporary approaches leverage remote sensing technologies to map fuels, burn histories, and environmental controls at unprecedented scales, forming the analytical backbone for modern risk models. Land-cover mosaics at 30-meter resolution from sources such as CEC, combined with fine-resolution topography from SRTM, characterize fuel type, continuity patterns, and terrain-driven fire behavior. Gridded weather and water-balance fields from Daymet and NOAA wind products, alongside EDDI drought indices, provide climate forcings and fuel-moisture proxies essential for risk assessment [24–26,35,42]. The LANDFIRE program integrates these diverse data streams into standardized fuel layers, enabling consistent, simulation-ready inputs across regions [36]. While these products offer comprehensive coverage and methodological consistency, they face inherent limitations in spatial resolution and update frequency. These constraints can obscure neighborhood-scale variability, recent disturbance patterns, and rapid moisture dynamics, producing coarse risk contrasts that fail to capture meaningful local variation [33,36]. These trade-offs highlight the need for frameworks that maintain national coverage while incorporating decision-scale interpretability and locally responsive drivers to support actionable planning.

***Machine learning advances and persistent challenges.*** Recent developments in machine learning and deep learning have substantially improved wildfire prediction capabilities by capturing nonlinear, high-dimensional interactions among fuels, weather, and terrain. Random forests, support vector machines, and neural network architectures have demonstrated significant gains in predicting ignition probability, fire spread patterns, and burn severity [10–14]. However, this literature predominantly optimizes for accuracy at regional or national scales while providing limited attention to transparent attribution and policy-ready explanations, effectively creating black-box predictors that resist practical interpretation [6,7,10–14]. Furthermore, models trained in one biophysical regime frequently demonstrate weak transferability to other regions because wildfire dynamics emerge from highly localized processes governed by context-specific fuel conditions, topoclimatic patterns, and ignition pressures. Consequently, performance gains achieved in one region often fail to transfer without substantial recalibration [6,7,23]. This combination of limited granularity, interpretive opacity, and uncertain portability continues to impede the translation of sophisticated analytics into community-scale risk management.

***Integration of exposure and social vulnerability dimensions.*** A complementary research strand advances beyond hazard assessment to incorporate exposure and social vulnerability, transforming hazard maps into comprehensive risk frameworks. National initiatives including FEMA's National Risk Index and the Wildfire Risk to Communities datasets integrate fire likelihood and intensity metrics with built-environment exposure and representative structure characteristics, identifying locations where human and infrastructure assets concentrate within high-hazard zones [15,16]. Empirical research documents how rapid growth at the wildland-urban interface combines with spatial inequities in mobility, housing quality, and healthcare access to amplify wildfire impacts even when biophysical hazard levels remain comparable [5]. However, many vulnerability assessments remain spatially coarse or temporally static, rarely coupling with interpretable, local response curves that clarify which specific drivers elevate risk within neighborhoods and by what magnitude. This gap limits the development of targeted fuel treatments, building standards, and evacuation infrastructure investments calibrated to local threshold conditions.

***Global frameworks and the scope-specificity trade-off.*** Early-warning systems and global assessment frameworks extend situational awareness capabilities but underscore fundamental trade-offs between scope and specificity. Near-real-time indices such as EDDI, combined with gridded meteorological products from Daymet and NOAA, support daily to seasonal outlooks that track atmospheric dryness and wind regimes associated with rapid fire escalation [24–26]. At global scales, assessments frequently employ proxies such as KBDI combined with land-cover data to evaluate wildfire potential under climate change scenarios. While useful for macro-scale comparisons, these approaches prove too generalized for decision-scale planning applications [23]. Collectively, current tools offer broad coverage and improving predictive skill but lack interpretable, place-sensitive diagnosis of nonlinear thresholds in critical drivers such as vegetation composition and elevation bands. Practitioners require this granular information to implement effective zoning regulations, vegetation management programs, evacuation infrastructure upgrades, and post-event response operations.

***Point of departure.*** The present study introduces WildfireGenome to address these persistent limitations through a fundamentally different approach. The framework fuses seven federal wildfire indicators into a PCA-based composite label, providing a unified risk metric [16,43]. It employs Random Forests to learn county-level risk classes, achieving high-resolution prediction at decision-relevant scales [44]. Through SHAP and ICE/PDP analyses, the framework reveals county-specific nonlinearities and systematically audits transferability across diverse geographic contexts [45]. This approach adapts the genome-style analytical blueprint successfully validated for flood risk assessment through FloodGenome to address the wildfire domain's greater spatial heterogeneity [46]. By combining predictive accuracy with transparent driver explanations at decision scale, WildfireGenome enables targeted interventions including vegetation management at empirically derived thresholds, elevation-band zoning calibrated to local risk patterns, and climate-sensitive readiness protocols. These capabilities complement rather than replace existing hazard products, enhancing their operational utility for local decision-making.

# Method

Our approach captures various wildfire risk predictors in creation of WildfireGenome models. The predictor set systematically incorporates the three fundamental controls of wildfire behavior: fuels, topoclimate, and weather. Each variable demonstrates clear mechanistic links to ignition likelihood, spread rate, and fire intensity, with all data harmonized to H3 Level-8 decision scale for operational relevance. Vegetation composition comprises thirteen fractional land-cover classes at 30-meter resolution that encode critical fuel characteristics. These classes capture fuel type, vertical and horizontal continuity, and potential firebreaks through specific mechanisms. Needleleaf forests indicate crown-fire propensity, while shrub and grass coverage represents fast-curing fine fuels that enable rapid spread. Wetlands and barren land create natural breaks in fuel continuity, and cropland reflects both altered fuel availability and modified ignition pressures from agricultural activities. Topographic variables at 30-meter resolution include elevation, slope, and latitude, each modulating fire behavior through distinct pathways. These factors influence microclimate conditions, channel wind patterns, facilitate upslope preheating, and mark ecozone transitions that fundamentally restructure fuel moisture profiles and ladder-fuel arrangements. Climate and weather variables capture both preconditioning factors and immediate forcing conditions. Annual precipitation, maximum temperature, and vapor pressure at one-kilometer resolution shape baseline fuel moisture levels and overnight recovery potential. The Evaporative Demand Drought Index at four-kilometer resolution captures cumulative atmospheric dryness and rapid flash-drought stress that primes fuels for ignition. Wind speed at ten meters directly controls spread rate and ember transport distances, representing immediate fire behavior forcing.

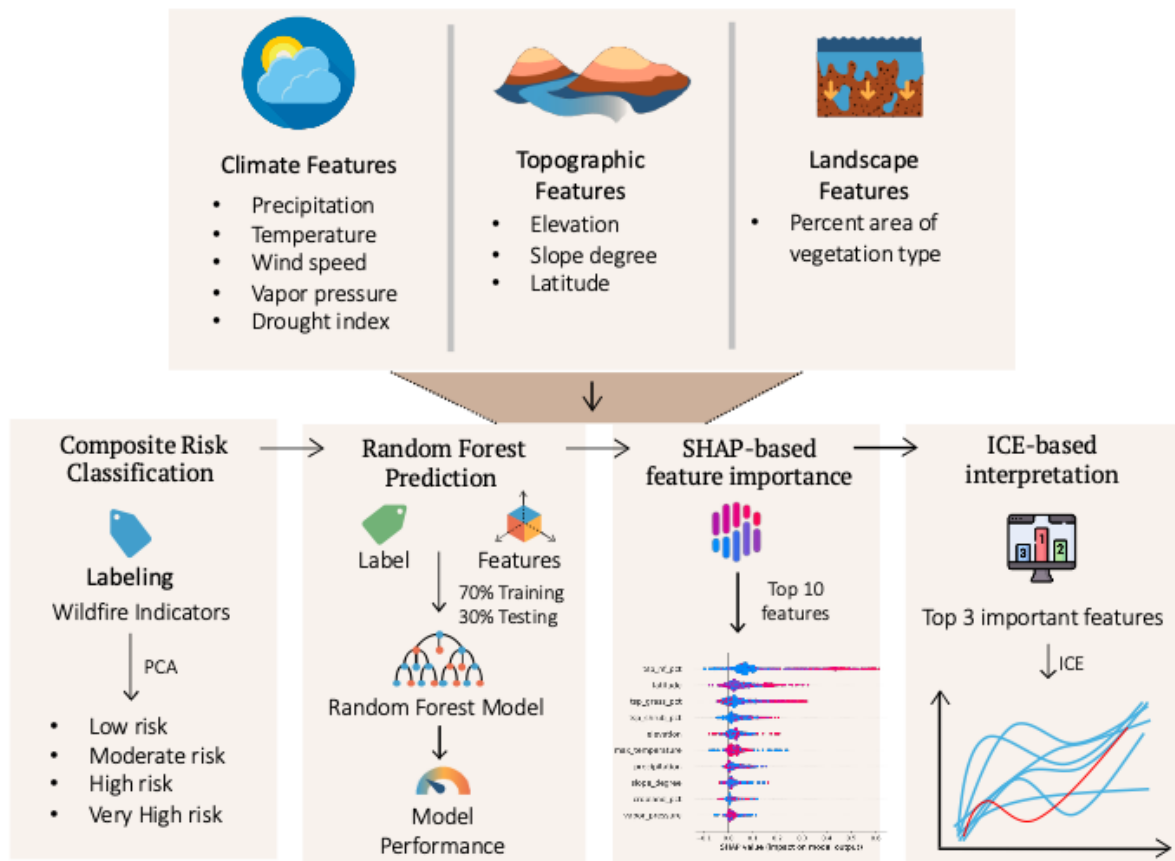

*Figure 1. WildfireGenome Framework for Interpretable Wildfire Risk Modeling. This diagram outlines the WildfireGenome modeling framework, which integrates environmental features and interpretable machine learning to assess spatial wildfire risk. The process begins with the extraction of climate, topographic, and landscape features, followed by the generation of composite wildfire risk labels through PCA-based classification. A Random Forest model is then trained to predict wildfire risk categories using the extracted features, with performance evaluated on a held-out test set. SHAP-based analysis identifies the top global predictors driving risk variation, while ICE-based interpretation reveals localized, nonlinear response patterns for the most influential features. This multi-tiered approach enables both robust risk prediction and fine-grained explanation, supporting adaptive land use, vegetation management, and migration planning under increasing wildfire threats.*

## Data description and processing

Table 1 summarizes the predictor groups and associated variables used in this study, including their original spatial resolutions and temporal coverage. The study area encompasses seven counties across four U.S. states: Butte, Sonoma, and El Dorado (California); Jackson (Oregon); Boulder and Larimer (Colorado); and Coconino (Arizona). All counties are classified as having high or very high wildfire risk according to the National Risk Index (NRI) [15]. To enable spatially consistent analysis, all datasets were aggregated to H3 Level 8 hexagonal grid cells within county boundaries. County selection was guided by several criteria: (1) representation of diverse ecoregions and fuel types—including Mediterranean California, the Pacific Northwest, the Rocky Mountains, and the Southwest Plateau; (2) coverage of environmental gradients in aridity, elevation, and vegetation; and (3) completeness and compatibility across the geospatial data products used, facilitating robust cross-county model transferability and generalization.

*Table 1. Summary of feature groups and variables*

| Feature groups | variable | description | resolution | year |
|---|---|---|---|---|
| Climate | precipitation | annual average of total precipitation | 1km | 2020 |
| | temperature | annual average of maximum temperature | 1km | |
| | vapor pressure | annual average vapor pressure | | |
| | wind_speed | annual average of wind speed above 10 meters | 1.9 degrees | |
| | drought index | drought index | 4km | |
| Topographic | elevation | elevation | 30 m | 2000 |
| | slope degree | slope degree | | 2021 |
| | latitude | latitude | - | - |
| Vegetation cover | tsp_nf_pct | % of temperate/subpolar needleleaf forest | 30m | 2020 |
| | ts_ble_pct | % of tropical/subtropical broadleaf evergreen forest | | |
| | ts_bld_pct | % of tropical/subtropical broadleaf deciduous forest | | |
| | tsp_bld_pct | % of temperate/subtropical polar broadleaf deciduous forest | | |
| | mixed_forest_pct | % of mixed forest | | |
| | ts_shrub_pct | % of tropical/subtropical shrubland | | |
| | tsp_shrub_pct | % of temperate/subpolar shrubland | | |
| | ts_grass_pct | % of tropical/subtropical grassland | | |
| | tsp_grass_pct | % of temperate/subpolar grassland | | |
| | wetland_pct | % of wetland | | |
| | cropland_pct | % of cropland | | |
| | barrenland_pct | % of barren land | | |

**Wildfire risk components data**

We compiled seven wildfire risk indicators from official U.S. wildfire datasets [16]: Burn Probability (BP), the long-term annual chance of fire occurrence; Flame Length Exceedance Probabilities at 4 ft and 8 ft (FLEP4 and FLEP8), which estimate the conditional probability that flames exceed 4 or 8 feet if a fire occurs; Canopy Fuel Load (CFL), the average flame length in the direction of maximum spread; Wildfire Hazard Potential (WHP), an index reflecting the relative potential for difficult-to-control fires; Risk to Potential Structures (RPS), a composite metric combining fire likelihood and intensity with generalized exposure to a representative structure at each location; and Exposure, the proportion of housing in a grid cell at risk from direct flame contact or indirect effects (e.g., embers, house-to-house spread). These indicators span a mix of probabilities, proportions, and continuous intensity measures, each with distinct scales and distributions.

Because principal component analysis (PCA) and distance-based models are sensitive to variable scale, skewness, and boundary effects, we applied tailored transformations to normalize the input features. For probability-type indicators (BP, FLEP4, FLEP8), we used a logit transformation to reduce boundary effects and promote homoscedasticity:

$$x^* = ln\left(\frac{x}{1-x}\right)$$

For proportion-based variables like Exposure, which are bounded and often right-skewed, we applied an arcsine square-root transformation, commonly used for fractional data:

$$x^* = arcsin(sqrt(x))$$

For heavily right-skewed indicators with wide dynamic ranges (e.g., CFL, WHP, RPS), we used a log-transformation to stabilize variance and avoid dominance by large outliers:

$$x^* = ln(1+x)$$

All transformed variables were subsequently standardized using a RobustScaler (centered by the median and scaled by the interquartile range), which down-weights the influence of extreme values and ensures that no feature is disproportionately emphasized due to scale.

**Climate feature**

We incorporated five climate-related predictors: total precipitation, maximum temperature, vapor pressure, wind speed, and a drought index. These variables influence fuel moisture, curing, and fire behavior. Reduced precipitation dries grasses, litter, and fine woody fuels, while unusually wet periods can increase fine-fuel loads that later dry out and ignite easily [17]. Elevated temperatures raise evaporative demand and reduce overnight fuel moisture recovery [18]. Higher wind speeds enhance fire spread and increase the potential for ember transport across firelines [19, 20]. Lower vapor pressure—indicative of drier air—accelerates moisture loss from live and dead fuels [21]. The drought index captures cumulative dryness over multi-week to multi-month timescales; elevated values indicate stressed vegetation and increased ignition susceptibility [22, 23]. Collectively, these variables describe how wet fuels are, how quickly they dry, and how readily fire can spread.

Climate data for precipitation, maximum temperature, and vapor pressure were obtained from the Daymet Version 4 product [24], which provides daily surface weather data aggregated to annual summaries on a 1 km × 1 km grid across North America. Specifically, we used annual totals for precipitation and annual means for maximum temperature and vapor pressure. Wind speed data were retrieved from NOAA [25], representing annual surface wind conditions (at 10 meters height) on a 1.9° × 1.9° global grid. The Evaporative Demand Drought Index (EDDI) [26] was used as our drought metric, providing annual values at a 4 km resolution. EDDI is designed to detect anomalies in atmospheric evaporative demand, offering early-warning insights into emerging drought conditions. Its sensitivity to rapid changes in moisture demand makes it particularly useful for anticipating both flash droughts and prolonged fire-weather stress.

**Topographic feature**

Topographic features considered in this study include elevation, slope (degrees), and latitude, each of which influences wildfire behavior through its effects on fuel conditions and fire dynamics. Elevation affects local temperature, snow accumulation, and vegetation patterns: higher elevations tend to be cooler and wetter, with longer snow cover, which delays fuel drying and the onset of fire season; lower elevations are typically warmer and drier, with greater prevalence of grass and shrub fuels and higher human activity, increasing ignition likelihood and spread potential [27, 28]. Slope influences fire spread rate: fires move more rapidly upslope as rising heat preheats vegetation ahead of the flame front, while gentle slopes allow for slower fire progression and greater overnight moisture recovery; steep slopes and canyons may channel wind and intensify fire behavior [29, 30]. Latitude serves as a proxy for solar exposure and broad climatic gradients: lower latitudes are generally warmer with longer dry seasons, whereas higher latitudes have shorter growing seasons but can still experience significant summer drying [31, 32]. Latitude is also associated with changes in vegetation types, further affecting fuel availability and continuity [33, 34]. Together, these topographic variables shape patterns of fuel moisture, fire spread potential, and the likelihood of extreme fire behavior. Elevation data were retrieved at 30-meter resolution using the Python elevation package [35]. Slope values were derived from 30-meter Digital Elevation Models (DEMs) provided by the LANDFIRE data products [36].

**Vegetation cover feature**

Vegetation composition plays a central role in shaping wildfire behavior by determining the fuel type, structure, and spatial continuity, which collectively influence ignition likelihood, spread dynamics, and fire intensity [37]. For example, needleleaf and evergreen forests often support ladder fuels and dense canopies that increase the likelihood of crown fires under conducive conditions [38]. In contrast, shrublands and grasslands contribute fine, fast-drying fuels that promote rapid fire spread [39]. Wetlands, barren land disrupt fuel continuity and can act as natural firebreaks, whereas croplands alter both fuel availability and human ignition pressures [40, 41]. Capturing the proportional coverage of these land-cover types enables a more nuanced representation of the fuel mosaic within each landscape, improving our ability to model spatial variation in wildfire hazard.

For each county, we quantified landscape composition as the areal fraction (%) of each vegetation or land-cover class within the analysis units. The classification scheme included: temperate/subpolar needleleaf forest; tropical/subtropical broadleaf evergreen forest; tropical/subtropical broadleaf deciduous forest; temperate/subtropical deciduous forest; mixed forest; tropical/subtropical shrubland; temperate/subpolar shrubland; tropical/subtropical grassland; temperate/subpolar grassland; wetland; cropland; barren land; and The percentages of all classes sum to 100% for each unit, and each class-specific percent cover was included as an independent predictor. Vegetation cover data were obtained from the Commission for Environmental Cooperation (CEC) at 30-meter spatial resolution [42].

# Method

## PCA

Given the presence of seven wildfire risk indicators, we applied Principal Component Analysis (PCA) to reduce dimensionality and extract dominant patterns across these correlated variables [43]. PCA is an unsupervised linear transformation technique that projects the original input features onto a smaller set of orthogonal components—principal components (PCs)—which are ordered by the proportion of variance they explain. Each component is a weighted linear combination of the original variables, enabling a compact and interpretable summary of multivariate structure while minimizing information loss.

We performed PCA on the standardized wildfire indicator matrix, and retained the first two components (PC1 and PC2), which together explained over 85% of the total variance across the study area (Table S1). Because PCA eigenvectors are sign-indeterminate, we oriented each retained component such that higher component scores reflect greater wildfire risk. Specifically, we computed Pearson correlations between each component's scores (across H3 grid cells) and a reference risk metric—Risk to Potential Structures (RPS), or Wildfire Hazard Potential (WHP) where RPS was unavailable. If the correlation was negative, we multiplied both the component scores and loadings by −1. This sign alignment step does not affect the variance explained, ranking, or inferential validity of PCA; rather, it ensures consistency in interpretation across counties and components.

Following alignment, we constructed a composite wildfire risk index by computing a weighted average of PC1 and PC2 scores for each grid cell, where the weights were proportional to the explained variance ratio of each component. This composite score provides a unified and interpretable measure of spatial wildfire risk intensity. The risk score for each grid cell i is calculated as:

$$Composite\ risk\ score_i = z(PC1)_i \times w_1 + z(PC2)_i \times w_2,$$

$$where\ w_1 = \frac{EVR_{pc1}}{EVR_{pc1} + EVR_{pc2}}, w_2 = \frac{EVR_{pc2}}{EVR_{pc1} + EVR_{pc2}}$$

## Random Forest

Random Forest is a supervised ensemble learning algorithm that builds a collection of decision trees from bootstrapped subsets of the data and aggregates their predictions through majority voting for classification task. It is a widely used ensemble learning method that constructs multiple decision trees and aggregates their predictions to improve classification accuracy and robustness [44]. To predict wildfire risk from environmental and landscape features, we trained a Random Forest (RF) classifier using the composite PCA-based risk score as the response variable. The continuous PCA scores were classified into four ordinal risk classes—low, moderate, high, and very high—based on empirical quartile thresholds pooled across all grid cells for each county. This binning approach enables categorical risk interpretation consistent with operational fire risk ratings.

The input feature set included all standardized predictors listed in Table 1, covering climate conditions, topographic attributes, and vegetation composition. No additional feature engineering or dimensionality reduction was applied prior to model fitting. We implemented the RF classifier using Scikit-learn's RandomForestClassifier, specifying 500 trees and using the Gini impurity criterion. Models were trained separately for each county, allowing the classifier to learn localized patterns in the feature–risk relationship. Model performance was evaluated via stratified 70/30 train-test splits within each county. Evaluation metrics included overall classification accuracy, macro-averaged F1 score and QWK, with additional analyses presented in the Results.

**SHAP**

To interpret the feature contributions learned by the Random Forest model, we used SHAP (SHapley Additive exPlanations), a unified framework for model interpretability grounded in cooperative game theory [45]. SHAP assigns each feature a Shapley value that quantifies its marginal contribution to the model's output, considering all possible feature combinations. These values represent a theoretically sound and consistent way of attributing predictions to individual input variables, offering model-wide interpretability.

In the context of wildfire risk modeling, SHAP enables us to move beyond black-box predictions and quantify how specific features—such as topographic attributes, vegetation composition, and climate variables—drive model outputs across space and time. This is particularly important for ensuring the transparency and actionability of machine learning insights in applied risk assessment.

We implemented SHAP using the TreeExplainer module from the SHAP Python package, which provides an exact and computationally efficient solution for tree-based models such as Random Forests. SHAP values were computed for all grid cells in the test data. To aggregate across multi-class outputs, we focused on expected SHAP values, which weight class-specific SHAP contributions by predicted class probabilities. This yields a single value per feature per instance, approximating its effect on the expected risk level. We then calculated global feature importance as the mean absolute SHAP value across all grid cells. To visualize these effects, we generated SHAP summary plots, including bar plots to rank features by importance and beeswarm plots to reveal the distribution and directionality of feature impacts.

To efficiently scale SHAP analysis across counties and grid cells, we used memory-mapped arrays for SHAP value storage and performed batch-wise computation to minimize memory footprint. This approach provides robust, interpretable explanations of how different landscape and climate predictors influence the spatial distribution of predicted wildfire risk.

**Individual Conditional Expectation (ICE) and Partial Dependence Analysis**

To visualize and interpret the feature-wise impact on predicted wildfire risk for individual observations, we employed Individual Conditional Expectation (ICE) plots in conjunction with

Partial Dependence Plots (PDP). For each county, we first identified the top three most influential features using the global SHAP importance scores derived from expected-level SHAP values. These top features were selected from a ranked list of feature contributions, previously computed and saved for each county.

We then used Partial Dependence Display from_estimator() from the scikit-learn library to generate ICE and PDP curves for each of the top three features, focusing on their marginal effect on the calibrated predicted probability of the "Very High" risk class. ICE curves were drawn for each test sample to capture individualized, nonlinear responses, while the average PDP curve (computed as the mean of the ICE curves) was overlaid and rendered in red for clarity. This approach complements the global SHAP analysis by revealing localized patterns such as thresholds, saturation effects, and non-monotonic responses that may be obscured in aggregate explanations. Together, the ICE and PDP visualizations provide a more nuanced understanding of how critical features affect extreme wildfire risk classifications.

# Results

## Spatial Patterns of Composite Wildfire Risk

Composite wildfire risk scores, generated through principal component analysis (PCA) of seven wildfire indicators, reveal clear spatial heterogeneity across the seven study counties. Each grid cell was classified into quantile-based risk categories to enable within-county comparison. In Butte, El Dorado, and Sonoma Counties (California), higher risk scores are concentrated in interior and peripheral areas, with lower scores observed near the central developed zones. Jackson County (Oregon) shows a north–south gradient, with moderate to high risk more common in the northern and central areas. In Coconino County (Arizona), elevated risk appears in the central and southeastern regions. Boulder and Larimer Counties (Colorado) display localized clusters of higher risk in their western portions. These spatial patterns reflect differences in the multivariate structure of the fire risk indicators and are not influenced by external predictors. As such, the maps provide a data-driven summary of internal variation in modeled wildfire risk, serving as a foundation for further interpretation and feature-level attribution.

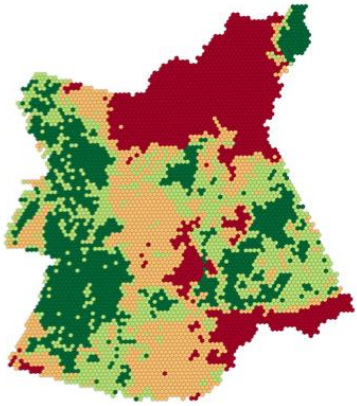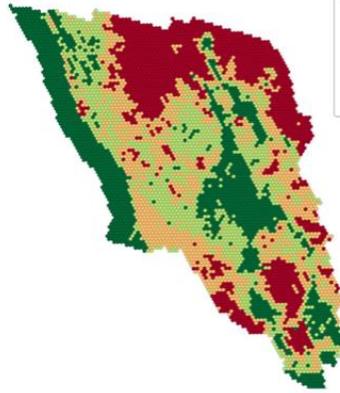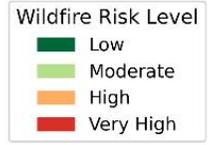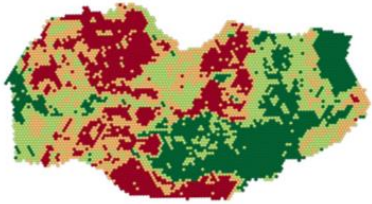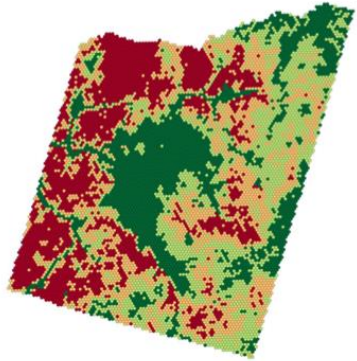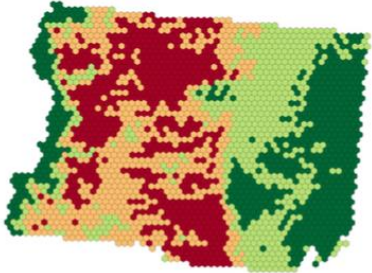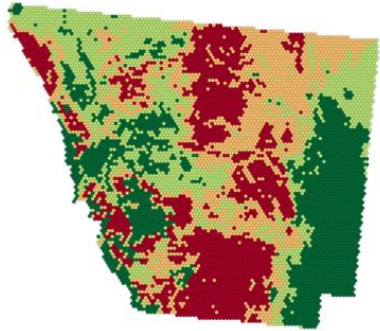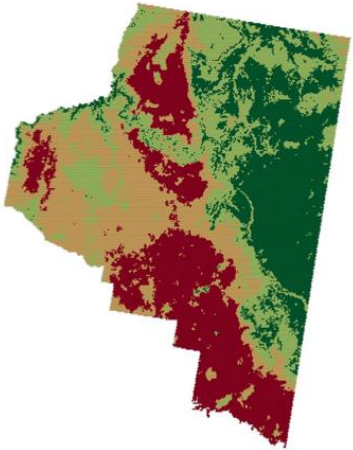

*Figure 2 Spatial Patterns of Composite Wildfire Risk Across Study Counties. Composite wildfire risk levels by H3 hexagonal grid cells for the seven study counties: Butte, Sonoma, and El Dorado (California); Jackson (Oregon); Boulder and Larimer (Colorado); and Coconino (Arizona). Risk categories (Low, Moderate, High, Very High) were assigned using quantile classification of composite PCA scores derived from seven wildfire hazard indicators. The maps reveal substantial spatial heterogeneity within and across counties, offering a comparative, data-driven snapshot of relative wildfire hazard without incorporating environmental covariates.*

## Random Forest performance

We trained a Random Forest classifier to predict quantile-based wildfire risk categories derived from the PCA composite index using all predictor groups—landscape, climate, vegetation, and topography. The dataset for each county was randomly split into 70% for training and 30% for testing. Within the training set, 5-fold cross-validation was applied to tune hyperparameters and evaluate model robustness. Because wildfire risk levels were assigned using equal-frequency quantiles, class distributions were balanced, reducing concerns about class imbalance that often affect multi-class classification.

*Table 2. Random Forest Performance Metrics by County*

| County | Accuracy | Macro-F1 | QWK |
| --- | --- | --- | --- |
| Butte County | 0.796 | 0.795 | 0.908 |
| Sonoma County | 0.779 | 0.78 | 0.893 |
| El Dorado County | 0.775 | 0.775 | 0.875 |
| Jackson County | 0.755 | 0.757 | 0.883 |
| Boulder County | 0.83 | 0.831 | 0.932 |
| Larimer County | 0.812 | 0.812 | 0.911 |
| Coconino County | 0.878 | 0.878 | 0.951 |

The random forest classifiers yielded consistently strong performance across all seven counties, with accuracy, macro-F1, and Quadratic Weighted Kappa (QWK) values ranging from 0.755 to 0.878 (Table 2). Coconino County achieved the highest scores across all three metrics (accuracy: 0.878, macro-F1: 0.878, QWK: 0.951), followed closely by Boulder and Larimer Counties, indicating robust generalization across varying landscape and hazard profiles. Even in counties with more heterogeneous landscapes—such as Jackson and El Dorado—the models maintained solid classification accuracy.

Confusion matrices (Figure 3) further illustrate model reliability by visualizing classification performance across wildfire risk levels. Most counties show strong diagonal dominance, indicating accurate predictions. Errors were primarily concentrated between adjacent risk categories (e.g., "Moderate" vs. "High"), which is expected in ordinal classification problems. Misclassification between non-adjacent levels (e.g., "Low" vs. "Very High") was rare across all counties, reinforcing that the models preserved the ordinal structure of wildfire risk. Notably, Coconino and Boulder exhibited particularly clean matrices with minimal off-diagonal dispersion, suggesting better class separability. In contrast, counties like Jackson and Sonoma showed slightly higher dispersion around the middle risk levels, reflecting more classification ambiguity in transition zones.

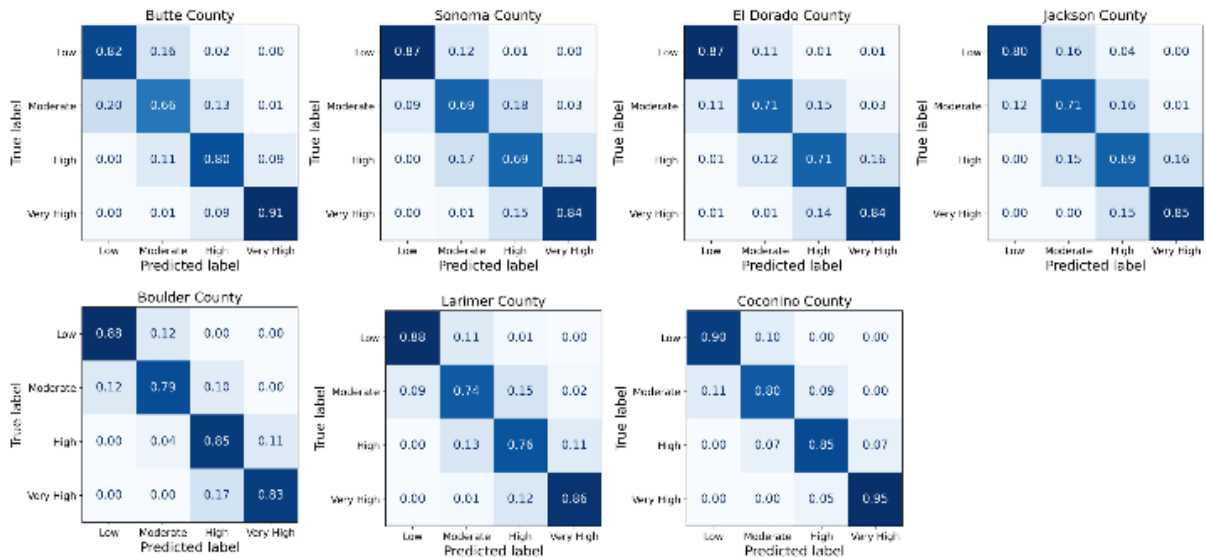

*Figure 3. Confusion Matrices for County-Level Wildfire Risk Classification.* *Confusion matrices summarize the predictive performance of the random forest classifier across seven counties, showing agreement between predicted and observed wildfire risk categories. Each cell indicates the proportion of grid cells classified into each true–predicted risk level pair. Most misclassifications occur between adjacent classes, indicating the model's ability to preserve relative ordering even when predictions are imperfect.*

## Model transferability

While the Random Forest model demonstrated strong predictive performance within each county—achieving high accuracy, macro-F1 scores, and ordinal agreement (QWK)—these results reflect conditions where training and testing data are drawn from the same spatial context. However, real-world applications often require models to operate in regions where no ground-truth wildfire risk labels are available. This raises a key question: To what extent do models trained in one county generalize to other counties with different landscape and climatic characteristics? To address this, we systematically evaluate cross-county model transferability by training on one county and testing on the remaining ones, thereby assessing the robustness and spatial portability of our learned risk patterns.

Results are summarized in Figure 4, which presents three heatmaps of transfer performance using different metrics: Balanced Accuracy, Macro-F1 and QWK. Each cell in the heatmap reflects the performance when the model is trained on the row county and tested on the column county. Diagonal elements represent within-county performance and serve as benchmarks.

The diagonal cells consistently exhibit high values across all metrics, confirming robust model performance when trained and tested within the same county. However, off-diagonal values vary substantially. Notably, Boulder and Larimer counties exhibit relatively high mutual transferability

(e.g., QWK = 0.85 and 0.65), likely due to shared biophysical characteristics within the Colorado Front Range. In contrast, transfer performance between Butte County and most others is low or even negative in QWK (e.g., -0.24 from Butte to Boulder), suggesting model disagreement beyond chance and highlighting the uniqueness of local fire regimes.
n.

Cross-county performance heatmaps reveal a striking pattern: models demonstrate robust within-county performance but highly variable off-diagonal generalization. Counties sharing similar biophysical contexts transfer knowledge effectively, as seen in the Boulder-Larimer pairing achieving QWK scores of approximately 0.85 and 0.65 respectively. In contrast, ecologically dissimilar county pairs perform poorly, sometimes below chance levels, exemplified by Butte to Boulder transfers yielding QWK scores of approximately -0.24. These patterns reveal two fundamental insights. First, wildfire susceptibility emerges from locally specific interactions among fuels, topography, and topoclimate that resist clean transfer across different ecological regimes. Second, the construction of labels through county-specific quartiles of composite risk indices introduces structural challenges for model portability. From a scientific perspective, these findings demonstrate that even highly accurate one-size-fits-all wildfire models can produce misleading results when applied to new geographies, contrasting sharply with prior FloodGenome research where cross-regional portability proved substantially stronger. For practical implementation, agencies should treat imported models exclusively as preliminary screening tools and plan to refit interpretable local models before operational deployment. Despite these limitations, strategic value exists in leveraging transferability within ecoregional clusters such as Front Range counties, where shared models can accelerate initial risk assessment while local calibration proceeds.

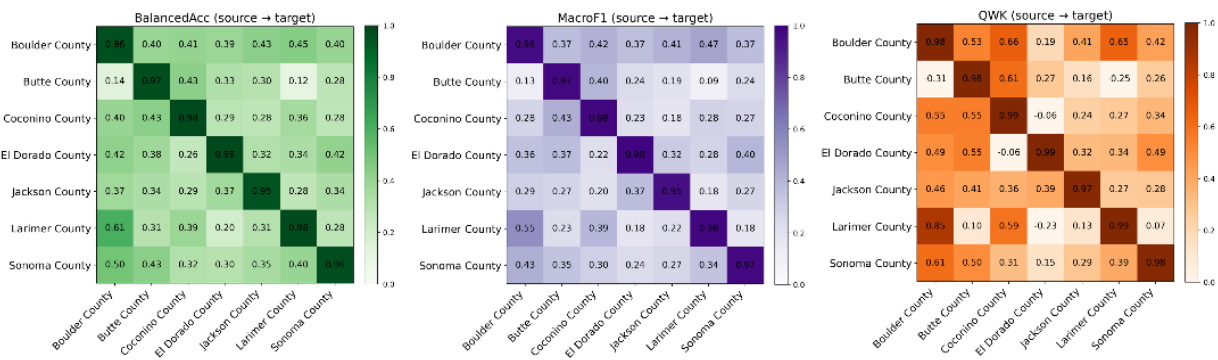

*Figure 4. Transferability of Random Forest models across counties evaluated using three performance metrics: Balanced Accuracy, Macro-F1 score, and Quadratic Weighted Kappa (QWK). Each cell represents the model performance when trained on the source county (rows) and tested on the target county (columns). Diagonal cells indicate within-county performance. Higher scores along the diagonal confirm strong local predictive performance, while varying off-diagonal values reveal substantial heterogeneity in cross-county generalizability. For instance, Boulder and Larimer counties exhibit mutual transferability, whereas Butte and El Dorado show limited predictive overlap with other counties. Negative or low QWK scores (e.g., Butte to Boulder) indicate model disagreement beyond chance, underscoring the importance of local context in wildfire risk modeling.*

# Localized Risk Responses and Implications for Migration Planning

To gain insight into the factors contributing to very high wildfire risk predictions, we employed SHAP values to quantify feature importance across all counties. Beeswarm plots of expected-level SHAP values (Figure 5) reveal both consistent and heterogeneous patterns in feature contributions across the seven counties. Across all counties, the percentage of needleforest vegetation (tsp_nf_pct) emerged as the most influential predictor, consistently associated with increased risk. This finding suggests that land cover types other than dense forest—such as grasslands, shrublands, or sparsely vegetated areas—may play a critical role in fueling or facilitating wildfire spread. For instance, high tsp_nf_pct values were especially prominent in Butte, El Dorado, Boulder, and Larimer Counties, where they had the highest average SHAP impacts; Elevation also consistently ranked among the top features in all counties, with higher elevations generally corresponding to increased predicted risk. This may reflect the interaction between topography, fuel types, and fire behavior in mountainous regions. Similarly, latitude was among the top contributors in 6 of 7 counties, likely capturing spatial gradients in vegetation, climate, and fire exposure that are not fully accounted for by the other predictors; Other physical and climatic variables exhibited more region-specific importance. Slope degree was a dominant feature in Sonoma and Jackson Counties, reflecting steeper terrain's association with more severe fire behavior. In contrast, meteorological drivers such as maximum temperature and vapor pressure were more influential in eastern counties such as Larimer and Coconino, indicating a stronger climatic signal in those regions; Several vegetation-specific features such as tsp_grass_pct and tsp_shrub_pct also showed moderate importance, particularly in Boulder and Butte Counties. Notably, lower values of precipitation and vapor pressure were frequently associated with increased predicted risk, highlighting the role of atmospheric dryness and drought conditions in exacerbating fire potential; Together, these results suggest that while some wildfire risk drivers are universal—such as land cover and elevation—others are highly context-dependent, shaped by regional biophysical and climatic conditions. This underscores the importance of localized modeling and interpretation when developing risk mitigation strategies.

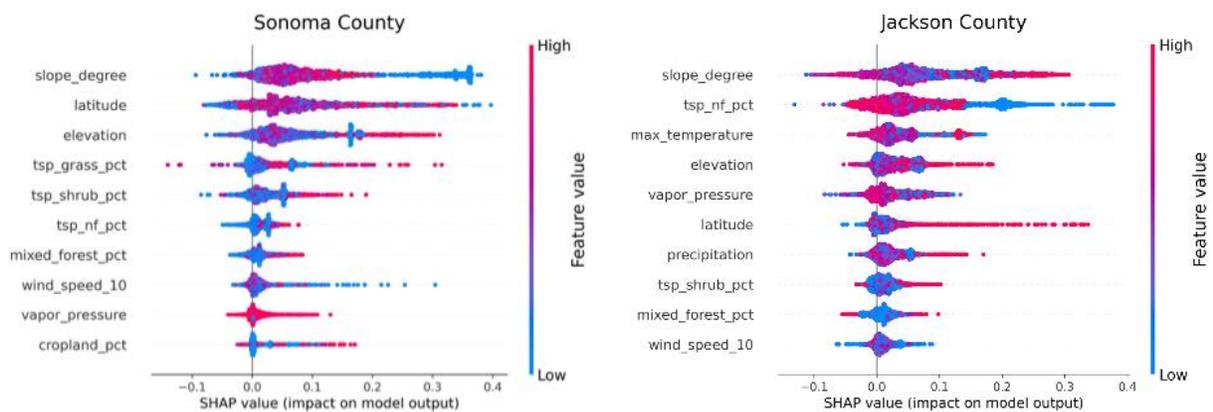

*Figure 5. Beeswarm plots illustrate the direction and magnitude of the top 10 features contributing to predicted wildfire risk in two representative counties (Sonoma County, Jackson County). Each point represents a test sample, with color indicating the feature value (red = high, blue = low). Horizontal spread reflects feature impact on model predictions relative to expected risk level. Feature importance and effect*

*direction vary across counties, highlighting the spatial heterogeneity in wildfire risk drivers. See Supplementary **Figure S1** for SHAP beeswarm plots from the remaining counties.*

While the SHAP summary plots provide a global ranking of feature importance and directionality, they do not capture the full extent of heterogeneous or nonlinear effects present at the individual level. To address this limitation and complement the global analysis, we examined Individual Conditional Expectation (ICE) and Partial Dependence (PDP) plots for the top three most influential features in each county, focusing specifically on their marginal effect on the predicted probability of "Very High" wildfire risk (Figure 6). These individualized plots offer fine-grained insights into how predicted risk responds to variation in each feature across test samples, revealing patterns that are often masked in aggregate summaries.

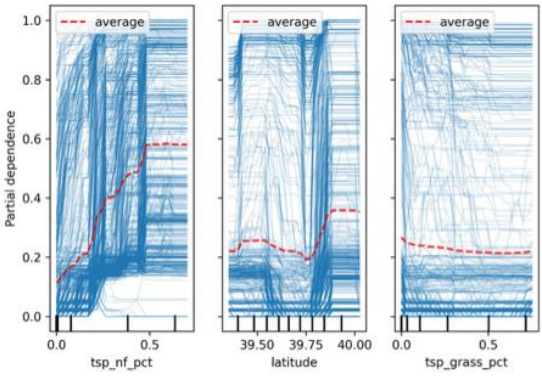
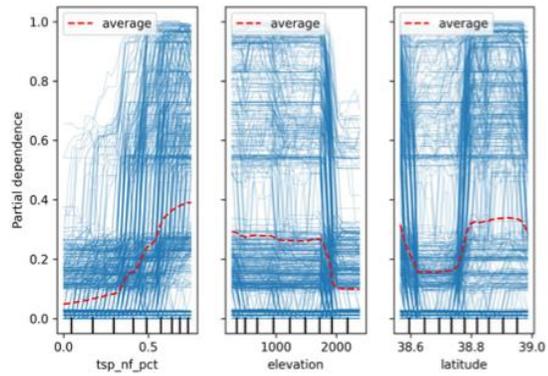
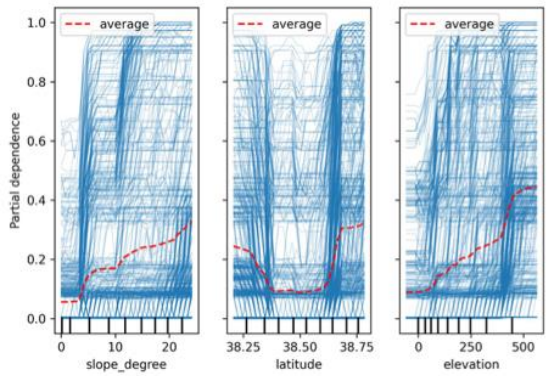
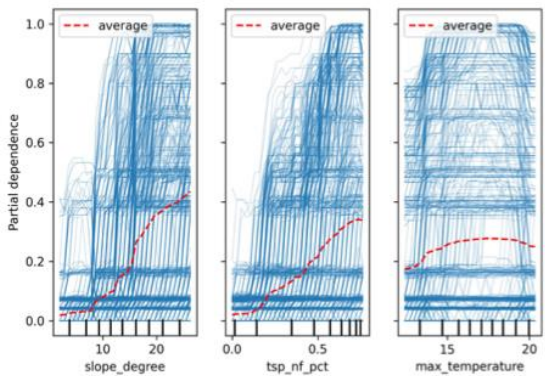
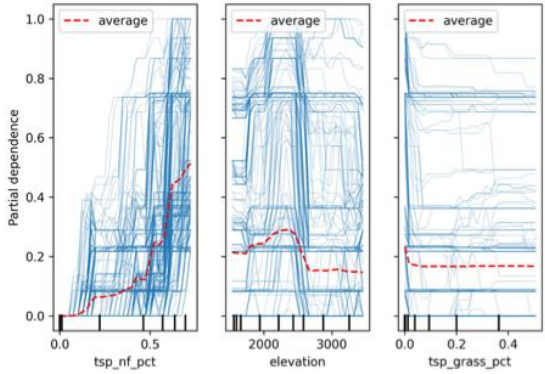
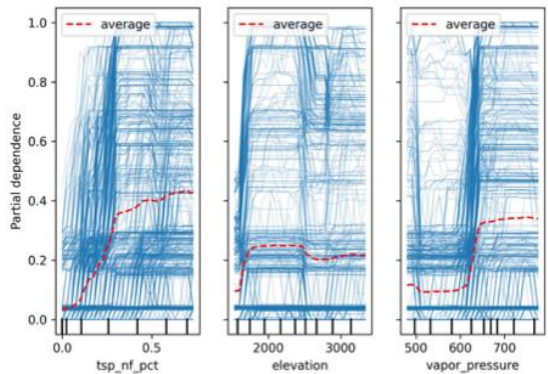
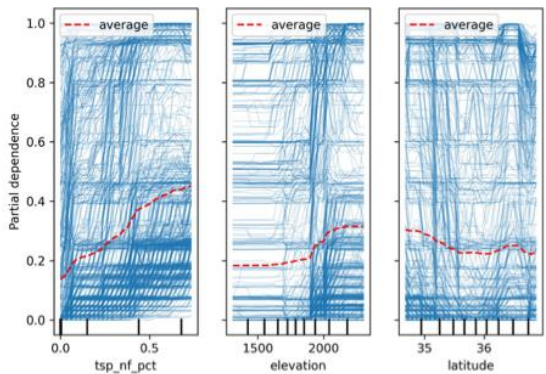

*Figure 6. ICE and PDP Plots of Top Predictors Across Seven Counties. Individual Conditional Expectation (ICE) and Partial Dependence (PDP) plots illustrating the marginal effects of the top three most influential features on the predicted probability of the "Very High" wildfire risk class across seven counties. For each county, the top predictors were selected based on expected-level SHAP importance scores. ICE curves (blue lines) capture the individualized responses of test samples, while the red dashed lines show the corresponding PDP curves, representing the average marginal effect. Forest cover—especially the percentage of temperate/subpolar needleleaf forest (tsp_nf_pct)—emerges as a consistently strong driver of wildfire risk across all counties. Elevation, slope, vapor pressure, latitude, and other vegetation types such as temperate/subpolar shrubland and grassland (tsp_shrub_pct, tsp_grass_pct) exhibit nonlinear or threshold-like behaviors, indicating county-specific ecological and climatic influences on fire susceptibility. These insights highlight spatial heterogeneity in feature-risk relationships and support localized adaptation and mitigation planning.*

Several commonalities and spatial heterogeneities emerged across counties. Notably, the percentage of temperate/subpolar needleleaf forest (tsp_nf_pct) consistently demonstrated a strong positive association with elevated wildfire risk, often exhibiting threshold-like patterns where risk probability sharply increased beyond certain forest coverage levels (e.g., ~30–40%). This trend was evident in counties such as Butte, El Dorado, Larimer, and Coconino, underscoring the role of specific vegetation types in shaping fuel load and flammability. Elevation and latitude also played critical roles, frequently modulating the effects of other variables. For instance, in El Dorado and Boulder counties, wildfire risk increased with elevation up to a point and then plateaued, indicating a non-monotonic relationship likely driven by transitions in vegetation or climatic regimes across altitudinal bands. Latitude effects were more county-specific but commonly reflected regional climatic gradients that influence exposure. Other land cover types, such as temperate/subpolar grassland (tsp_grass_pct) and shrubland (tsp_shrub_pct), showed more mixed or even risk-reducing associations in certain counties. For example, in Butte and Boulder, higher grassland coverage was associated with lower or flat risk responses. These individualized response curves illuminate substantial variation in how wildfire risk responds to ecological and topographic gradients both across and within counties. Importantly, the ICE curves expose localized sensitivities and nonlinearities—such as saturation zones, inflection points, and step-change behaviors—that are not captured by global summary plots, making them critical for informing spatially targeted policy interventions.

Building on these individualized response patterns, we now explore how such insights can inform practical decision-making for climate-resilient migration and land-use adaptation. Counties exhibiting threshold behaviors—such as a sharp increase in fire risk beyond 40% tsp_nf_pct—may require targeted vegetation management policies, including thinning, understory removal, or forest diversification strategies, to mitigate risk in situ. In contrast, counties where increased vegetation does not significantly elevate risk (e.g., Sonoma or Coconino) may be more suitable for managed in-migration, provided that other resilience metrics (e.g., water security, evacuation infrastructure) are satisfied. The elevation-dependent patterns also suggest that mid-elevation zones should be considered transitional or high-risk areas, whereas lowland or stable plateau regions with consistent or declining ICE curves may offer climate refuge potential.

In operational terms, ICE-informed migration planning can enable policymakers to classify regions not solely by historical fire records or uniform hazard zones, but rather by latent ecological responsiveness to key features. For example, settlements could be prioritized in zones where

tsp_grass_pct is moderately high and tsp_nf_pct remains below critical thresholds, as indicated by stable or declining ICE responses. Additionally, planning tools can incorporate ICE-informed thresholds to dynamically update land eligibility for development, conservation, or managed retreat. Risk communication strategies may also be tailored to populations living in "sensitive response zones" to encourage mitigation compliance or relocation with clearer justification. Ultimately, ICE-based insights help transition from reactive to anticipatory risk governance—guiding not only where people should migrate, but also how to adaptively reshape land systems to accommodate those movements while minimizing cumulative fire exposure.

In sum, Global SHAP analyses combined with localized ICE and PDP curves reveal a nuanced landscape of universal drivers operating alongside context-contingent sensitivities. Needleleaf forest share and elevation consistently emerge as top predictors across all counties, yet their effect sizes and functional relationships vary substantially by location. Slope demonstrates particularly strong influence in Sonoma and Jackson counties, while temperature and vapor pressure, indicators of atmospheric dryness, carry greater predictive weight in Larimer and Coconino. The ICE curves reveal critical nonlinear thresholds that inform management decisions. Risk increases sharply when needleleaf coverage exceeds approximately 30 to 40 percent across multiple counties, while elevation effects exhibit non-monotonic patterns likely reflecting ecotone transitions between vegetation zones. These findings carry significant scientific and practical implications. Scientifically, they argue for place-based inference that extends beyond simple feature ranking to encompass full response functions. Practically, they enable precisely targeted interventions: forest thinning and understory management become priorities where needleleaf thresholds are exceeded; zoning revisions and infrastructure hardening warrant consideration in mid-elevation bands where risk plateaus at high levels; and monitoring systems should focus on vapor pressure anomalies as leading indicators in dryness-sensitive counties. The critical insight is that effective wildfire risk management requires understanding not merely which features matter, but how they function within specific local contexts. This granular, place-specific information forms the foundation for credible county-level adaptation strategies and operational planning.

## Concluding Remarks

This study introduces the WildfireGenome, a high-resolution and interpretable modeling framework designed to characterize spatial wildfire risk and identify its localized drivers across ecologically diverse U.S. counties. Wildfire risk is commonly assessed through coarse hazard maps, fire-behavior simulators, or empirical indices that summarize long-term fire records. While recent machine learning and deep learning models have improved prediction accuracy, they typically operate at regional or national scales and function as black boxes. These models provide little insight into the fine-scale, place-specific interactions among vegetation, topography, and climate that drive actual risk at the local level where land-use, infrastructure, and migration decisions occur. Current approaches have three critical shortcomings: they rarely integrate multiple official risk components into a single, analytically coherent framework; they seldom expose nonlinear thresholds in key drivers that inform policy decisions, such as vegetation composition tipping points; and they infrequently test whether models and their explanations transfer effectively across distinct geographies. This creates a persistent gap between high-level hazard products and the interpretable, decision-scale evidence required for targeted adaptation strategies.

The WildfireGenome framework directly addresses these limitations through three key innovations. First, it fuses seven federal wildfire indicators into a composite, sign-aligned PCA risk label at H3-grid resolution. Second, it learns local risk classes using Random Forests. Third, it opens the model through SHAP for global attribution and ICE/PDP for individual response curves, revealing county-specific, nonlinear sensitivities that are actionable for zoning, vegetation management, and climate-resilient migration planning. These sensitivities include critical thresholds for factors such as needleleaf forest coverage and elevation. The study rigorously evaluates both within-county performance and cross-county transferability across seven ecologically diverse counties, empirically demonstrating where model portability fails and why place-based modeling is essential. This approach advances the field from coarse, opaque, and monolithic risk mapping to a scalable yet context-sensitive analytics framework that combines predictive skill with policy-ready explanations at the decision scale.

The models demonstrate strong within-county performance, with accuracy ranging from 0.755 to 0.878, comparable Macro-F1 scores, and Quadratic Weighted Kappa values reaching 0.951. Confusion matrices reveal that errors predominantly occur between adjacent risk classes, indicating high ordinal fidelity. This performance profile enables reliable in-county prioritization and triage strategies. Agencies can confidently designate "Very High" risk cells as first-priority targets and treat surrounding "High" risk cells as buffer zones, aligning operational response with model reliability while minimizing the risk of extreme false negatives. The rare occurrence of "Low" to "Very High" misclassifications further supports this approach. Since risk labels represent county-relative quartiles rather than absolute measures, agencies should apply these maps exclusively for ranking actions within individual counties rather than making direct cross-county comparisons.

Cross-county transferability exhibits marked heterogeneity in performance. Models transfer effectively between ecologically similar regions, such as the Front Range counties of Boulder and Larimer, maintaining reasonable predictive power. However, transfers between dissimilar ecological contexts yield poor results, sometimes performing below chance levels, as demonstrated by Butte to Boulder transfers producing QWK scores of approximately -0.24. These findings mandate specific operational protocols. Imported models should function solely as preliminary screening tools, with agencies computing local composite labels and refitting county-specific models before any high-stakes deployment. When time constraints preclude model refitting, practitioners should select ecoregional donor models from within the same mountain belt or ecological zone and implement conservative operating rules. These should include expanded buffer zones and lowered escalation thresholds to account for elevated transfer uncertainty.

The SHAP and ICE analyses provide actionable intelligence by revealing not only which factors drive risk but also how these factors behave within specific local contexts. Two universal drivers, needleleaf forest share and elevation, consistently rank as primary predictors across all counties. However, their response functions exhibit substantial local variation, with critical thresholds frequently occurring at approximately 30 to 40 percent needleleaf coverage and non-monotonic patterns in elevation effects. These insights enable precisely targeted interventions across multiple management domains.

For vegetation management, agencies should prioritize thinning and understory treatments in areas where ICE curves indicate risk surges beyond local needleleaf thresholds. In land-use planning, stricter zoning regulations and infrastructure hardening become warranted in elevation bands where risk plateaus at elevated levels. Counties where slope emerges as a dominant driver, particularly Sonoma and Jackson, require focused investment in evacuation route upgrades and egress infrastructure. In climate-sensitive counties such as Larimer and Coconino, where atmospheric dryness carries greater predictive weight, operational readiness and resource prepositioning should align with vapor pressure and temperature anomalies as well as drought signals. The ICE-derived response curves enable definition of specific operational triggers. Agencies can escalate patrols, inspections, and fuel treatments in cells where forecast atmospheric dryness coincides with vegetation coverage beyond identified risk thresholds. For long-term planning, stable-response zones identified through flat or declining ICE curves represent favorable locations for in-migration or development, while threshold-sensitive zones warrant constrained investment. This approach transforms wildfire risk management from uniform hazard zoning to county-specific operational frameworks anchored in transparent driver rankings and empirically derived response functions.

This study advances wildfire risk science through several key innovations. First, it fuses seven federal risk components into a sign-aligned, PCA-based composite label at the decision scale, creating a unified risk metric. Second, it employs interpretable machine learning to expose county-specific, nonlinear driver thresholds that shape local risk patterns. The study goes beyond demonstrating strong within-county accuracy by conducting a systematic audit of cross-county transferability across seven ecologically diverse counties. This analysis reveals precisely when and why model portability fails, providing critical insights for applied risk assessment. Through SHAP and ICE/PDP analyses, the framework identifies both universal drivers, such as needleleaf forest coverage and elevation, and place-contingent sensitivities that vary by location. The combination of the composite-labeling scheme with paired global and local interpretability provides a transparent, reproducible template that addresses three critical gaps in current literature: fragmented hazard indicators, opaque models, and untested generalizability assumptions.

The framework produces policy-ready, grid-level risk surfaces and actionable response curves that directly support zoning decisions, vegetation and fuel treatment planning, evacuation and egress infrastructure upgrades, and automated post-wildfire damage assessment using multiview ground-view images. The system identifies threshold-sensitive H3 cells where risk increases sharply beyond specific needleleaf coverage levels, enabling agencies and utilities to strategically pre-position baseline street-level image collection. Following a wildfire event, these organizations can deploy rapid re-imaging through vehicle-mounted, body-worn, or other ground-view cameras to automate change detection and structure-level damage grading. The interpretable drivers inform three critical operational decisions: where to concentrate imaging passes, how densely to sample viewpoints, and which neighborhoods require priority triage. This approach reduces assessment latency and optimizes resource allocation. Furthermore, the explicit transferability results guide model adaptation and data acquisition strategies when fires occur in new counties, effectively aligning pre-event risk intelligence with post-event operational workflows.

The WildfireGenome framework complements rather than replaces existing wildfire assessment tools by addressing three critical gaps between coarse hazard products and operational decision-making requirements. First, the framework harmonizes seven federal risk components—BP, FLEP4/8, CFL, WHP, RPS, and Exposure—into a sign-aligned PCA composite, then predicts quartiled risk classifications at H3 decision scale. This approach produces within-county rankings that enable direct operational action, moving beyond broad regional hazard zones to provide actionable intelligence at the resolution where planning decisions occur. Second, the framework transforms opaque model outputs into transparent policy guidance. Unlike black-box machine learning approaches or fire-behavior simulators that lack interpretable attributions, WildfireGenome combines Random Forest models with SHAP and ICE/PDP analyses to reveal local, nonlinear response curves. These analyses identify specific thresholds, such as critical needleleaf forest coverage levels and elevation bands, that convert risk maps into concrete policy triggers. These triggers directly inform zoning decisions, targeted fuel and vegetation treatments, retrofit prioritization, and evacuation infrastructure upgrades. Third, the comprehensive transferability audit provides practical guidance for model reuse across jurisdictions. Agencies gain explicit knowledge about when models trained in other counties serve as credible screening proxies, particularly for ecologically similar regions, and when rapid local model refitting becomes essential. This operational guidance remains absent from most current risk assessment products.

These integrated capabilities enhance current wildfire risk assessment methods by making them more targeted, explainable, and portable at the precise scale where land-use planning and emergency management decisions are implemented. The framework thus bridges the persistent gap between regional risk assessment and local operational needs, providing decision-makers with scientifically grounded, actionable intelligence tailored to their specific geographic contexts.

# Acknowledgements

This material is based in part upon work supported by the National Science Foundation under CRISP 2.0 Type 2 No. 1832662 grant. Any opinions, findings, conclusions, or recommendations expressed in this material are those of the authors and do not necessarily reflect the views of the National Science Foundation.

# Code availability

The code that supports the findings of this study is available from the corresponding author upon request.

# Supplementary information

Table S1. Explained Variance Ratios (EVR) of PC1/PC2

| State | County | EVR-PC1(%) | EVR-PC2(%) | Cumulative-EVR(%) |
|---|---|---|---|---|
| CA | Butte County | 78.7 | 11.7 | 90.4 |
| | Sonoma County | 82.81 | 10.9 | 93.71 |
| | El Dorado County | 73.7 | 17.3 | 91 |
| OR | Jackson County | 74.5 | 18.6 | 93.1 |
| CO | Boulder County | 83.7 | 8.1 | 91.8 |
| | Larimer County | 72.8 | 14 | 86.8 |
| AZ | Coconino County | 73.5 | 17.8 | 91.3 |

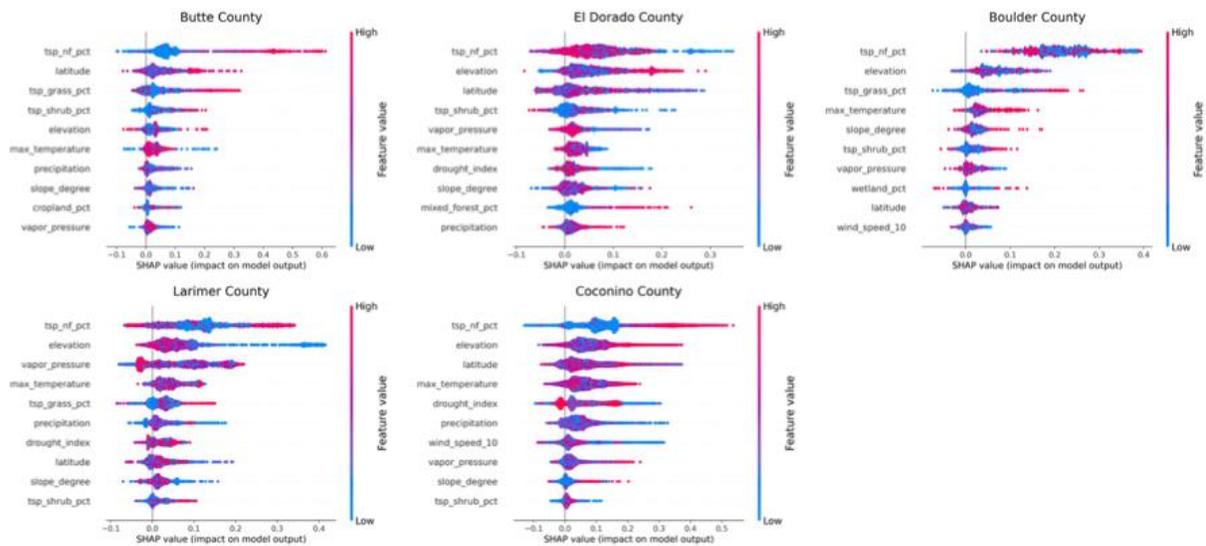

Figure S1. *Beeswarm plots illustrate the direction and magnitude of the top 10 features contributing to predicted wildfire risk in other 5 counties*